\documentclass[10pt,twocolumn,letterpaper]{article}

\usepackage{iccv}              

\usepackage{booktabs}
\usepackage{pgfplotstable}
\usepackage{siunitx} 
\usepackage{tabularx} 
\usepackage{float}
\usepackage{xcolor}
\usepackage{multirow}
\usepackage{graphicx}
\usepackage{subcaption} 

\newcommand{\vx}{\mathbf{x}}
\newcommand{\vy}{\mathbf{y}}
\newcommand{\vz}{\mathbf{z}}
\newcommand{\vh}{\mathbf{h}}
\newcommand{\vw}{\mathbf{w}}
\newcommand{\vu}{\mathbf{u}}
\newcommand{\vv}{\mathbf{v}}
\newcommand{\vS}{\mathbf{s}}

\newcommand{\vH}{\mathcal{H}}
\newcommand{\vW}{\mathcal{W}}

\newcommand{\better}[1]{\textcolor{Green}{\bf+#1}}

\newcommand{\mypar}[1]{\vspace{0.35\baselineskip}\noindent\textbf{#1}.\,}

\usepackage{bbm}

\usepackage{amsmath}
\usepackage{amssymb}
\usepackage{graphicx,scalerel}
\newcommand\wh[1]{\hstretch{1.6}{\hat{\hstretch{.625}{#1}}}}
\newcommand\wt[1]{\hstretch{1.6}{\tilde{\hstretch{.625}{#1}}}}

\newcommand{\ppm}[1]{\,\scriptsize{$\pm$#1}}

\newcommand{\phz}{\phantom{0}}

\definecolor{iccvblue}{rgb}{0.21,0.49,0.74}
\usepackage[pagebackref,breaklinks,colorlinks,allcolors=iccvblue]{hyperref}

\def\paperID{12646} 
\def\confName{ICCV}
\def\confYear{2025}

\title{CTA: Cross-Task Alignment for Better Test Time Training\\
\textit{\normalsize A preprint}}

\author{Samuel Barbeau$^{1}$\thanks{Correspondence to \texttt{samuel.barbeau.2@ens.etsmtl.ca}} 
\quad
Pedram Fekri$^2$
\quad
David Osowiechi$^{1}$
\quad
Ali Bahri$^1$
\\
Moslem Yazdanpanah$^{1}$
\quad
Masih Aminbeidokhti$^1$
\quad
Christian Desrosiers$^1$ \\[0.5em]
$^1$ÉTS Montréal \quad $^2$Concordia University
}

\begin{document}
\maketitle
\begin{abstract}

Deep learning models have demonstrated exceptional performance across a wide range of computer vision tasks. However, their performance often degrades significantly when faced with distribution shifts, such as domain or dataset changes. Test-Time Training (TTT) has emerged as an effective method to enhance model robustness by incorporating an auxiliary unsupervised task during training and leveraging it for model updates at test time. In this work, we introduce CTA (Cross-Task Alignment), a novel approach for improving TTT. Unlike existing TTT methods, CTA does not require a specialized model architecture and instead takes inspiration from the success of multi-modal contrastive learning to align a supervised encoder with a self-supervised one. This process enforces alignment between the learned representations of both models, thereby mitigating the risk of gradient interference, preserving the intrinsic robustness of self-supervised learning and enabling more semantically meaningful updates at test-time. Experimental results demonstrate substantial improvements in robustness and generalization over the state-of-the-art on several benchmark datasets.

\end{abstract}    
\section{Introduction}
\label{sec:intro}

One of the key challenges in deep learning (DL) research is ensuring that models generalize effectively to new data. Typically, models are trained on a source dataset and evaluated on previously unseen images from the same distribution. While advanced DL models achieve remarkable performance on various benchmarks, the assumption of domain invariance between training and test data is often unrealistic in real-world applications. Consequently, DL models struggle with distribution shifts, limiting their robustness and practical deployment~\cite{torralba2011unbiased, miller2021accuracy}.

To address this challenge, two major research directions have emerged: domain generalization (DG) and unsupervised domain adaptation (UDA). DG approaches focus on training models with an inherent ability to generalize across diverse domains. However, a key limitation of these methods is their reliance on large, diverse training datasets, which are often impractical to obtain~\cite{zhou2021domain, cha2021swad}. Additionally, these techniques may struggle to perform well in domains that significantly differ from those encountered during training.

Conversely, UDA aims for greater generalizability without explicitly anticipating distribution shifts, instead adapting the model dynamically using either test-time adaptation (TTA) \cite{venkateswara2017deep, liang2020we} or test-time training (TTT) \cite{sun2020test, hakim2023clust3}. TTA methods adapt without access to the initial source training step or test sample labels. Our work focuses on TTT, which relaxes TTA's strict constraints by allowing source training to be designed in a way that facilitates adaptation. During training, TTT approaches employ a multi-task learning strategy, where the model simultaneously learns an auxiliary self-supervised or unsupervised task alongside the main objective, sharing encoder features across tasks. At test time, the model relies solely on the auxiliary task to update the shared encoder using a loss previously optimized for the source domain.

Current state-of-the-art TTT methods rely on custom model architectures to incorporate an auxiliary task. This is often done by adding new network modules to a common backbone in the form of branches \cite{rec-ttt,NCTTT2024,hakim2023clust3,ttt++}. However, the objective of TTT is to ensure practical applicability in real-world scenarios where implementing a new model architecture is often infeasible. More importantly, the multi-task learning approach of standard TTT models has been showed to suffer from negative transfer during joint training \cite{ruder2017overviewmultitasklearningdeep} where conflicting tasks lead to gradient interference \cite{yu2020gradientsurgerymultitasklearning}. As a result, \cite{ttt++} observed that the self-supervised updates performed by the auxiliary branch at test-time can be in conflict with the main objective. Under severe distribution shift, this lack of alignment can cause the auxiliary branch to update the feature representations of the backbone further away, leading to suboptimal performance. 

\begin{figure*}[t]
  \centering
  \includegraphics[width=.95\textwidth]{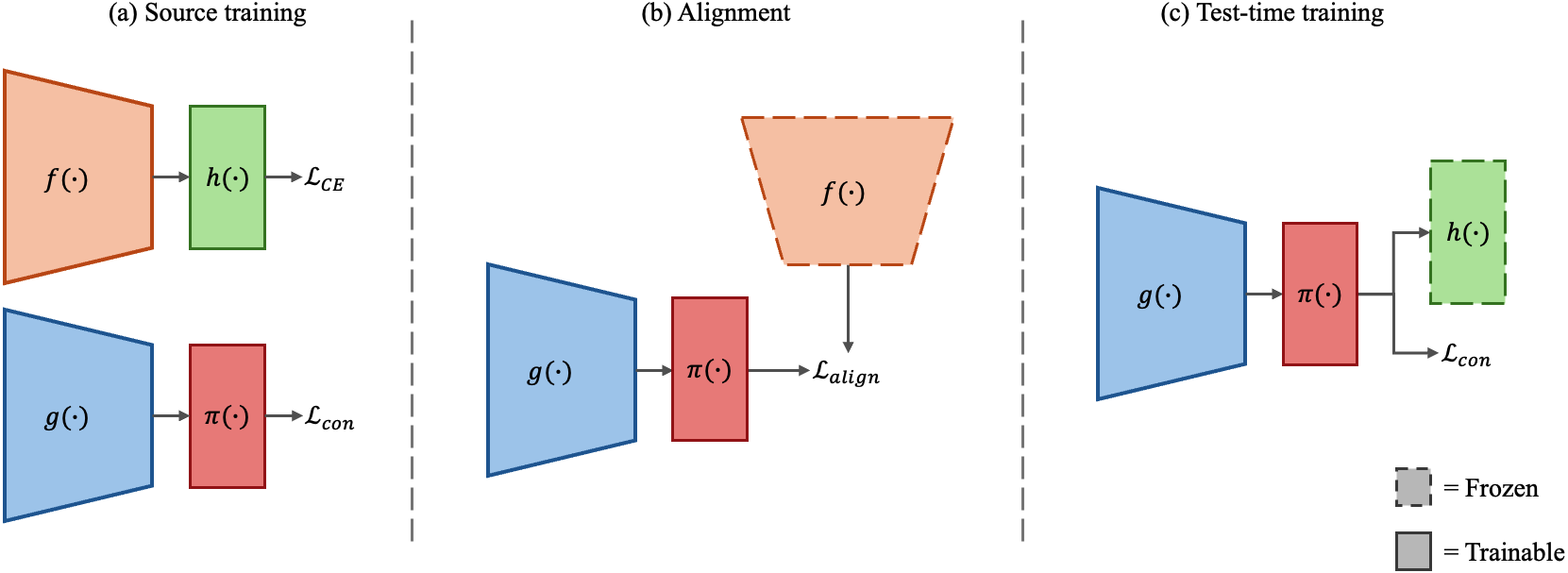}
  \caption{\textbf{Overview of CTA}. During the source training stage \textbf{(a)}, two distinct models \(h(f(\cdot))\) and \(g(\cdot)\) are trained using a common dataset but with different objectives, namely cross-entropy \(\mathcal{L}_{\text{CE}}\) and contrastive learning \(\mathcal{L}_{\text{con}}\) respectively. During the alignment stage \textbf{(b)}, the self-supervised model \(\pi(g(\cdot))\) is trained using source data to match the feature distribution of the frozen supervised encoder \(f(\cdot)\) using the contrastive loss \(\mathcal{L}_{\text{align}}\) described in \cref{eq:lalign}. At test-time \textbf{(c)}, we make use of the self-supervised model \(\pi(g(\cdot))\) to extract feature vectors for the frozen classifier \(h(\cdot)\). We update \(\pi(g(\cdot))\) using the same loss function it has seen during source training, namely \(\mathcal{L}_{\text{con}}\).}
  \label{fig:overview}
\end{figure*}

Motivated by these limitations, we propose CTA, an architecture-agnostic method that performs Cross-Task Alignment between two encoders to enforce alignment of learned representations within the same latent space. To this end, our method only requires that the supervised encoder be duplicated and fine-tuned using a self-supervised loss. Inspired by multi-modal contrastive learning \cite{radford2021learning}, the new self-supervised encoder can then be aligned with the frozen supervised encoder, leading to a joint latent space. After alignment, we show that the self-supervised model can be used as a feature extractor for the main task and updated using the self-supervised loss. By allowing distinct models to specialize on their respective tasks and distilling the decision boundary of the supervised objective via self-supervised learning (SSL), we avoid negative transfer and preserve the full generalizability of the self-supervised backbone. Additionally, our method differs from previous works by updating the entire model, eliminating the need to specify an \textit{extract layer} hyperparameter. An overview of our method is illustrated in \cref{fig:overview}.

The key contributions of our work are summarized as follows:
\begin{itemize}
\setlength\itemsep{.25em}
    \item We introduce CTA, a novel TTT approach that aligns representations between two encoders. Our method then utilizes only the self-supervised encoder for both the main and auxiliary tasks without degradation of performance. By distilling the supervised decision boundary via a SSL task, CTA mitigates conflicts between both objectives and preserves intrinsic robustness, leading to improved robustness under distribution shifts.
    \item Unlike current state-of-the-art methods, CTA is architecture-agnostic, allowing seamless integration with pretrained models without modifying their structure. Rather than training from scratch, CTA enables efficient adaptation through fine-tuning, making it more practical for real-world deployment.    
    \item Evaluated across diverse and challenging TTA scenarios with varying domain shifts, CTA demonstrates substantial performance improvements compared to previous methods.
\end{itemize}

\section{Related Works}\label{sec:related-works}

\mypar{Test-Time Adaptation (TTA)} In visual recognition tasks, distribution shifts between training and test data can significantly degrade model performance \cite{saenko2010adapting}. To address this challenge, recent approaches have focused on dynamically adapting models at test time to better handle new data. Unlike domain generalization \cite{cha2021swad, zhou2022domain}, which aims to train a robust source model that remains fixed during testing, test-time adaptation (TTA) allows models to be updated specifically for the target domain.

TTA operates without access to the original training data; instead, it adapts solely at test time using the pre-trained model. Various TTA methods have emerged in recent years. For example, PTBN \cite{nado2020evaluating} updates the BatchNorm layer statistics using the test batch, while TENT \cite{wang2020tent} minimizes prediction entropy to improve adaptation. More recently, TIPI \cite{nguyen2023tipi} identifies transformations that approximate domain shifts and trains the model to be invariant to them. Other methods have explored TTA for vision-language models. For instance, TPT~\cite{tpt} introduces Prompt Tuning, while CLIPArTT~\cite{clipartt} and WATT~\cite{watt} leverage pseudo-labeling to adapt the layer normalization parameters of the vision encoder.

\mypar{Test-Time Training (TTT)} Unlike Test-Time Adaptation, TTT techniques have access to the source data during initial training but not at test time. A secondary self-supervised task is jointly trained alongside the main learning objective. This paradigm was first introduced in TTT \cite{sun2020test}, where the auxiliary task involved recovering a randomly applied 90-degree rotation. During inference, adaptation occurs by updating only the parameters related to the secondary task.

TTT-MAE \cite{gandelsman2022test} employs transformers as the backbone for supervised training, using a masked autoencoder architecture for self-supervised reconstruction. At test time, the network adapts by reconstructing masked images, refining the shared feature extractor. TTTFlow \cite{osowiechi2023tttflow} applies normalizing flows on top of a pre-trained network, mapping features to a simple multivariate Gaussian distribution; adaptation is guided by the log-likelihood of this distribution. ClusT3 \cite{hakim2023clust3} introduces an unsupervised clustering task that maximizes Mutual Information between features and cluster assignments, ensuring consistency across different domains. TTT++ \cite{liu2021ttt++} builds upon the TTT framework by incorporating a contrastive approach, where two augmented versions of the same image serve as positive pairs, while augmented versions of different images act as negative pairs. Additionally, it employs batch-queue decoupling to regularize adaptation with smaller batch sizes. NC-TTT \cite{osowiechi2024nc} introduces a contrastive method that leverages the synthetic generation of noisy feature maps. The recently proposed ReC-TTT \cite{colussi2024rec} method leverages cross-reconstruction between a frozen encoder and two trainable encoders as auxiliary pre-training task and adaptation at test time.

Our approach differs from prior methods as it eliminates the need for a custom model architecture, making it compatible with pretrained weights and more realistic for real-world applications. Additionally, we directly tackle the problem of conflicting updates between the main and auxiliary tasks by introducing an alignment loss between a supervised and self-supervised encoder to enforce a joint representation space. This differs from TTT++ \cite{ttt++}, which uses saved statistics from the source domain as an additional alignment loss during test-time training.

\section{Methodology}
\label{sec:methodology}

Let $P_s(X, Y)$ represent the joint distribution of the source domain, where $\vx_s\!\in\!\mathcal{X}_s$ and $\vy_s\!\in\!\mathcal{Y}_s$ denote the input images and their corresponding labels, respectively. Similarly, let $P_t(X, Y)$ represent the joint distribution of the target domain, with $\vx_t\!\in\!\mathcal{X}_t$ and $\vy_t\!\in\!\mathcal{Y}_t$ as input images and labels in the target environment. A key assumption in test-time training (TTT) is that the conditional distribution of input given labels may differ between the source and target domains, i.e., $P(X_s | Y_s) \neq P(X_t | Y_t)$, while the label space remains consistent ($Y_s = Y_t$).

Given a model $F\!:\mathcal{X} \rightarrow \mathcal{Y}$ trained on labeled source data $(\vx, \vy)\!\in\mathcal{X}_s\!\times\!\mathcal{Y}_s$, the objective of TTT is to adapt this model to unlabeled target domain examples $\vx_t\!\in\!\mathcal{X}_t$ during inference. This adaptation is achieved by optimizing the model parameters using a self-supervised loss function computed on the target data. 

Our method aligns a supervised model $h(f(\cdot))$, where $h(\cdot)$ represents  a classifier, with a self-supervised version of itself $\pi(g(\cdot))$, where $\pi(\cdot)$ denotes a linear layer. To do this, we first train two architecturally identical models on a supervised and self-supervised task respectively within the source domain $\vx_s$. In real-world applications, a pretrained model could be duplicated to create a self-supervised copy. Once both models are trained, their representation spaces are aligned using a contrastive loss, again in the source domain $\vx_s$. Following this alignment step, we show that by using the self-supervised model as a feature extractor for the supervised task, we can adapt the encoder to large domain shifts at test-time. As a result, our method demonstrates improved robustness compared to baseline.


\begin{table*}[h]
\centering
\begin{tabular}{@{}lccccccc@{}}
    \toprule
     & & \multicolumn{2}{c}{\textbf{CIFAR10-C}} & \multicolumn{2}{c}{\textbf{CIFAR100-C}} & \multicolumn{2}{c}{\textbf{TinyImageNet-C}} \\
    \cmidrule(lr){3-4} \cmidrule(lr){5-6} \cmidrule(lr){7-8}
    \textbf{Encoder} & \textbf{Temperature} $\tau$ & \textbf{Source} & \textbf{Target} & \textbf{Source} & \textbf{Target} & \textbf{Source} & \textbf{Target} \\
    \midrule
    \multirow{1}{*}{Supervised} & — & 95.81 & 65.99 & 80.54 & 42.93 & \textbf{73.69} & \textbf{25.45} \\
    \midrule
    \multirow{6}{*}{Aligned Self-Supervised} 
    & 0.010  & 96.34 & 69.59 & 78.98 & 43.75 & 69.94 & 22.86 \\
    & 0.025  & 96.40 & 71.05 & 79.73 & 44.21 & 70.45 & 23.31 \\
    & 0.050  & 96.43 & 69.65 & 81.18 & 44.78 & 71.30 & 23.02 \\
    & 0.100  & 96.65 & 70.82 & 81.91 & 44.84 & 72.45 & 23.99 \\
    & 0.250  & 96.83 & 70.48 & 82.79 & 46.38 & 72.89 & 24.58 \\
    & 0.500  & \textbf{96.86} & \textbf{71.73} & \textbf{82.96} & \textbf{46.66} & 72.77 & 24.69 \\
    \bottomrule
  \end{tabular}
\caption{Classification results ($\%$) of the classifier \(h(\cdot)\) using the aligned self-supervised model \(\pi(g(\cdot))\) as a feature extractor with different datasets. We report results for CIFAR10-C, CIFAR100-C, and TinyImageNet-C across different temperatures \(\tau\).}
\label{tab:alignment_results}
\end{table*}

\subsection{Choice of self-supervised task}
\label{sec:self_supervised_task}

To isolate the effects of our alignment strategy, we employ the same self-supervised task as \cite{ttt++}, specifically SimCLR \cite{SimCLR}. This choice is deliberate as it ensures that any improvements observed are attributable to the proposed alignment method rather than differences in the underlying self-supervised learning approach. However, CTA is agnostic to the choice of self-supervised task and we leave it to future work to explore more efficient alternatives.

Given a mini-batch of $B$ images, we apply the same transformations as \cite{ttt++} to generate two augmented views per image. Each augmentation is mapped to a feature representation $\vz_i = g(\vx_i)$. This feature vector is then passed through a linear layer producing $\vh_i = \pi(\vz_i)$. While linear layers in the context of contrastive learning are often used to reduce dimensionality, we preserve the input dimension to ensure consistency with the output of the supervised encoder. The two augmentations originating from the same image form a positive pair $\langle \wh{\vh}_i, \wt{\vh}_i\rangle$, while all other pairs within the batch are treated as negatives. Denoting as $\wh{\vH} = \{\wh{\vh}_i\}_{i=1}^B$ and $\wt{\vH} = \{\wt{\vh}_i\}_{i=1}^B$ the two sets of augmented samples for the mini-batch, the contrastive learning objective is defined as:
\begin{equation}
    \mathcal{L}_{con} \, = \, -\sum_{i=1}^B\log \frac{\exp\big(\text{sim}(\wh{\vh}_i, \wt{\vh}_i)/\tau\big)}{\sum\limits_{\vh \in\, \wh{\vH} \cup \wt{\vH}} \!\!\mathbb{I}_{(\vh \neq \wh{\vh}_i)} \cdot \exp\big(\text{sim}(\wh{\vh}_i, \vh)/\tau\big)}
    \label{eq:lcon}
\end{equation}
where $\tau$ is a temperature parameter that scales the similarity scores and $\mathbb{I}_{\mathit{cond}}$ is the indicator function which equal to $1$ if $\mathit{cond}$ is true, else to $0$. The similarity measure used is cosine similarity, defined as:
\begin{equation}
    \text{sim}(\vu, \vv) \, = \, \frac{\vu^\top \vv}{\|\vu\| \|\vv\|}
    \label{eq:cosinesim}
\end{equation}


\subsection{Choice of supervised task}
\label{sec:supervised_task}

The choice of self-supevised task is dictated by the downstream application of the model. In our case, we evaluate our proposed method on common TTT classification benchmarks. Consequently, the supervised model $h(f(\cdot))$ is trained using the cross-entropy loss:
\begin{equation}
\mathcal{L}_{\text{CE}} \, = \, \frac{1}{B} \sum_{s=1}^B \sum_{k=1}^C y_{sk} \log \hat{y}_{sk}
\label{eq:lce}
\end{equation}
where $C$ is the number of classes and $\hat{y}_{sk} = \big[h(f(\vx_s))\big]_k$


\subsection{Pre-training the encoders}
\label{sec:pretraining_encoders}

The first stage of CTA is the pretraining of the supervised and self-supervised models on their respective tasks. Although CTA only requires that the output dimension of both models be identical to ensure compatibility with the alignment phase, we employ two architecturally identical models. These are trained independently from each other on their respective tasks but using the same source dataset $\vx_s$.

The self-supervised model is trained using the contrastive learning loss $\mathcal{L}_{con}$ described in \cref{sec:self_supervised_task} while the supervised model is trained using the cross entropy loss $\mathcal{L}_{\text{CE}}$ described in \cref{sec:supervised_task}.


\subsection{Aligning the encoders}
\label{sec:aligning_encoders}

Following the pre-training of the encoders, we align the feature distribution of both tasks by applying contrastive learning on the triplet $\langle \wh{\vh}_i, \wt{\vh}_i, \vw_i \rangle$ where $\wh{\vh}_i$ and $\wt{\vh}_i$ are the encoded augmentation pair from the image described in \cref{sec:self_supervised_task} and $\vw_i=h(f(\vx_i))$ is the output of the supervised encoder.

As the cosine similarity cannot be applied to vector triplets, we instead employ a cross-encoder contrastive loss with  $\langle \wh{\vh}_i, \vw_i\rangle$ and $\langle \wt{\vh}_i, \vw_i\rangle$ as positive pairs: 
\begin{equation}
\begin{aligned}
    \mathcal{L}_{align} &\, = \, -\sum_{i=1}^B\log \frac{\exp\big(\text{sim}(\wh{\vh}_i, \vw_i)/\tau\big)}{\!\!\sum\limits_{\vS \,\in \wh{\vH} \cup \vW} \!\!\mathbb{I}_{(\vS \neq \wh{\vh}_i)} \cdot \exp\big(\text{sim}(\wh{\vh}_i, \vS)/\tau\big)} \\
    & \ -\sum_{i=1}^B\log \frac{\exp\big(\text{sim}(\wt{\vh}_i, \vw_i)/\tau\big)}{\!\!\sum\limits_{\vS \,\in \wt{\vH} \cup \vW} \!\!\mathbb{I}_{(\vS \neq \wt{\vh}_i)} \cdot \exp\big(\text{sim}(\wt{\vh}_i, \vS)/\tau\big)}
\end{aligned} 
\label{eq:lalign}
\end{equation}
where $\vW = \{\vw_i\}_{i=1}^B$.

We investigated the impact of allowing gradient updates in the supervised encoder during the alignment process. To prevent deviation from its primary task, we added the cross-entropy loss from the frozen classifier \(h(f(\cdot))\) to the alignment loss. While this approach can yield good results, we observed that freezing the supervised encoder leads to more stable feature alignment. This setup resembles a student-teacher framework \cite{hinton2015distillingknowledgeneuralnetwork}, wherein the self-supervised encoder learns to match the feature distribution of the supervised encoder while simultaneously optimizing its own objective.

\begin{table*}[t]
\centering
\resizebox{.95\linewidth}{!}{
  \setlength{\tabcolsep}{10pt}
  \begin{tabular}{lcccccccc}
    \toprule
    \multirow[c]{2}{*}{\textbf{Corruption Type}} & \multirow[c]{2}{*}{\textbf{ResNet50}} & {\textbf{TTT++}} & {\textbf{ClusT3}}  & {\textbf{NC-TTT}} & {\textbf{ReC-TTT}} & \multirow[c]{2}{*}{\textbf{CTA-C} (Ours)}  & \multirow[c]{2}{*}{\textbf{CTA} (Ours)} \\[-3.5pt]
    & & \scriptsize{(NeurIPS21)} & \scriptsize{(ICCV23)} &  \scriptsize{(CVPR24)} &  \scriptsize{(WACV25)} & & \\
    \midrule
    Gaussian Noise & 26.86 & 75.87\ppm{5.05} & 75.81\ppm{2.62} & 75.24\ppm{0.12} & 71.97\ppm{1.18} & 76.42 & \textbf{83.09}\ppm{0.08} \\
    Shot Noise & 32.37 & 77.18\ppm{1.36} & 77.32\ppm{2.14} & 77.84\ppm{0.15} & 75.44\ppm{1.02} & 77.88 & \textbf{84.16}\ppm{0.12} \\
    Impulse Noise & 15.19 & 70.47\ppm{2.18} & 67.97\ppm{2.78} & 68.77\ppm{0.15} & 69.28\ppm{0.27} & 63.06 & \textbf{71.26}\ppm{0.22} \\
    Defocus Blur & 90.91 & 86.02\ppm{1.35} & 88.10\ppm{0.20} & 88.22\ppm{0.04} & 89.56\ppm{0.18} & 90.54 & \textbf{93.56}\ppm{0.06} \\
    Glass Blur & 57.22 & 69.98\ppm{1.62} & 60.47\ppm{1.72} & 70.19\ppm{0.18} & 69.38\ppm{0.73} & 68.39 & \textbf{75.84}\ppm{0.17} \\
    Motion Blur & 75.01 & 85.93\ppm{0.24} & 84.99\ppm{0.49} & 86.82\ppm{0.10} & 88.94\ppm{0.03} & 86.58 & \textbf{90.41}\ppm{0.06} \\
    Zoom Blur & 92.42 & 88.88\ppm{0.95} & 86.76\ppm{0.29} & 88.36\ppm{0.10} & 89.65\ppm{0.27} & 92.14 & \textbf{94.83}\ppm{0.02} \\
    Snow & 79.69 & 82.24\ppm{1.69} & 81.46\ppm{0.39} & 84.42\ppm{0.07} & 86.75\ppm{0.44} & 84.34 & \textbf{89.54}\ppm{0.03} \\
    Frost & 75.01 & 82.74\ppm{1.63} & 80.73\ppm{1.25} & 84.80\ppm{0.06} & 86.83\ppm{0.59} & 85.13 & \textbf{90.23}\ppm{0.03} \\
    Fog & 73.03 & 84.16\ppm{0.28} & 82.52\ppm{0.25} & 86.81\ppm{0.12} & 88.87\ppm{0.33} & 85.69 & \textbf{90.93}\ppm{0.03} \\
    Brightness & 91.02 & 89.97\ppm{1.20} & 91.52\ppm{0.24} & 92.52\ppm{0.04} & 94.03\ppm{0.24} & 93.13 & \textbf{95.12}\ppm{0.02} \\
    Contrast & 53.44 & 86.60\ppm{1.39} & 82.59\ppm{0.92} & 87.84\ppm{0.11} & 89.56\ppm{0.48} & 91.03 & \textbf{94.66}\ppm{0.06} \\
    Elastic Transform & 80.32 & 78.46\ppm{1.83} & 80.04\ppm{0.35} & 80.23\ppm{0.06} & 81.66\ppm{0.32} & 79.49 & \textbf{85.39}\ppm{0.06} \\
    Pixelate & 69.62 & 82.53\ppm{2.01} & 81.69\ppm{0.58} & 81.93\ppm{0.22} & 82.13\ppm{0.34} & 87.05 & \textbf{90.51}\ppm{0.08} \\
    JPEG Compression & 77.76 & 81.76\ppm{1.58} & 81.58\ppm{1.18} & 78.49\ppm{0.09} & 79.69\ppm{0.12} & \textbf{83.27} & 81.81\ppm{0.04} \\
    \midrule
    Average & 65.99 & 81.52 & 80.24 & 82.17 & 82.92 & 82.94 & \textbf{87.42} \\
    $\Delta$(Ours$-$Other) & \better{21.43} & \better{5.90}  & \better{7.19} & \better{5.26} & \better{4.51} & \better{4.48}
    & -- \\
    \bottomrule
  \end{tabular}
  }
  \caption{ Accuracy ($\%$) on CIFAR10-C dataset with level 5 corruption for CTA and recent state-of-the-art methods.}
  \label{tab:cifar10c_results}
\end{table*}


\subsection{Test-time training}

The objective of TTT is to adapt to a target domain $\vx_t$ relying exclusively on unlabeled examples. The alignment procedure described in \cref{sec:aligning_encoders} has the effect of instilling the decision boundary of the classifier $h(\cdot)$ in the feature distribution of the self-supervised model $\pi(g(\cdot))$ while preserving the contrastive objective the latter was originally optimized for. Consequently, we use $\pi(g(\cdot))$ as a feature extractor for the frozen classifier $h(\cdot)$, resulting in the test-time model $h(\pi(g(\cdot)))$.
To adapt to the target domain $\vx_t$, we apply the  contrastive learning objective $\mathcal{L}_{con}$ defined by \cref{eq:lcon} on the self-supervised model while keeping the classifier frozen. Unlike previous work \cite{hakim2023clust3,NCTTT2024,rec-ttt}, we update all parameters of the feature extractor, thereby eliminating the need for an \textit{update-layer} hyperparameter.
\section{Experimental Settings}
\label{sec:experiments}

\subsection{Datasets}
\label{sec:datasets}
Following previous work \cite{NCTTT2024, rec-ttt, sun19ttt}, we evaluate our proposed method on three publicly available datasets which simulate common image corruptions.

\mypar{Common image corruptions} To evaluate our method on common corruptions, we first train on CIFAR-10 (60,000 images, 10 classes, 32×32 resolution), CIFAR-100 (60,000 images, 100 classes, 32×32 resolution) \cite{CIFAR10}, and TinyImageNet (110,000 images, 200 classes, 64×64 resolution) \cite{TinyImagenet}. Following source training, we perform test-time training on the CIFAR-10-C, CIFAR-100-C, and TinyImageNet-C datasets \cite{hendrycks2019robustness}. Each image is synthetically augmented with 15 different types of corruption at 5 levels of severity. All experiments were conducted using the highest severity level (5).

\subsection{Implementation Details}
\label{sec:implementation_details}

Following previous work \cite{rec-ttt}, we use a ResNet50 \cite{He2015} pretrained on Imagenet \cite{imagenet} as a convolutional feature extractor, followed by a linear layer for both models. We use the same image transformations as \cite{ttt++, NCTTT2024, rec-ttt} for our supervised task and the same augmentations as \cite{ttt++} for our self-supervised task. The source training and alignment stages use the hyperparameter values described in \cref{tab:source_training}. Our test-time training employs the same setting as \cite{NCTTT2024}, namely a batch size of 128 and fixed learning rate of $1 \times 10^{-6}$. All experiments are run on a single NVIDIA GeForce RTX 3090 graphics card.

\begin{table}[h]
\centering
\resizebox{\columnwidth}{!}{
  \begin{tabular}{@{}lcc@{}}
    \toprule
    \textbf{Hyperparameter} & \textbf{Source training} & \textbf{Alignment} \\
    \midrule
    Epochs & 50 & 50 \\
    Batch size & 256 & 256 \\
    Optimizer & Adam & Adam \\
    Scheduler & Cosine annealing & Cosine annealing \\
    Starting learning rate & $5 \times 10^{-4}$ & $5 \times 10^{-4}$ \\
    Final learning rate & $1 \times 10^{-6}$ & $1 \times 10^{-6}$ \\
    Warmup epochs & 2 & 2 \\
    Temperature & 0.01 & 0.5 \\
    \bottomrule
  \end{tabular}
  }
  \caption{Source training and alignment settings}
  \label{tab:source_training}
\end{table}
\section{Results}
\label{sec:results}

\begin{table*}[h]
\centering
  \setlength{\tabcolsep}{10pt}
  \begin{tabular}{lccccc}
    \toprule
    \multirow[c]{2}{*}{\textbf{Corruption Type}} & \multirow[c]{2}{*}{\textbf{ResNet50}} & {\textbf{ClusT3}}  & {\textbf{NC-TTT}} & {\textbf{ReC-TTT}}  & \multirow[c]{2}{*}{\textbf{CTA} (Ours)} \\[-3.5pt]
    &  & \scriptsize{(ICCV23)} &  \scriptsize{(CVPR24)} &  \scriptsize{(WACV25)} &\\
    \midrule
    Gaussian Noise & \phz14.89 & 52.79 & 46.03 & 48.12 & \textbf{55.19}\\
    Shot Noise & \phz18.30 & 52.91 & 47.04 & 50.43 & \textbf{57.24} \\
    Impulse Noise & \phz8.95 & \textbf{45.54} & 41.53 & 45.29 & 41.57 \\
    Defocus Blur & 70.79 & 66.66 & 67.00 & 71.21 & \textbf{75.31} \\
    Glass Blur & 24.55 & \textbf{50.76} & 48.08 & 49.94 & 46.33 \\
    Motion Blur & 60.65 & 62.92 & 64.31 & 68.86 & \textbf{70.53} \\
    Zoom Blur & 74.21 & 65.42 & 66.24 & 69.91 & \textbf{77.34} \\
    Snow & 50.48 & 56.65 & 58.70 & 60.21 & \textbf{65.29} \\
    Frost & 45.39 & 56.91 & 58.55 & 60.16 & \textbf{66.83} \\
    Fog & 40.05 & 53.95 & 57.73 & 62.22 & \textbf{66.53} \\
    Brightness & 68.98 & 66.78 & 71.36 & 73.47 & \textbf{79.13} \\
    Contrast & 22.93 & 56.46 & 61.53 & 67.06 & \textbf{75.63} \\
    Elastic Transform & 52.60 & 59.07 & 60.25 & \textbf{62.37} & 59.13 \\
    Pixelate & 40.50 & 62.26 & 61.17 & 63.61 & \textbf{67.89} \\
    JPEG Compression & 50.61 & \textbf{59.34} & 55.69 & 57.05 & 55.30 \\
    \midrule
    Average & 42.93 & 57.89 & 57.68 & 60.66 & \textbf{63.95} \\
    $\Delta$(Ours$-$Other) & \better{21.02}  & \better{6.05} & \better{6.27} & \better{3.29} & --\\
    \bottomrule
  \end{tabular}
  \caption{Accuracy ($\%$) on CIFAR100-C dataset with level 5 corruption for CTA and recent state-of-the-art methods.}
  \label{tab:cifar100c_results}
\end{table*}

\subsection{Effectiveness of encoder alignment}
\label{sec:alignment_effectiveness}

Following the alignment phase, we evaluate its effectiveness by providing the classifier \(h(\cdot)\) with the features of the self-supervised model \(\pi(g(\cdot))\). \cref{tab:alignment_results} shows the alignment phase yields a self-supervised model \(\pi(g(\cdot))\) that outperforms the supervised encoder in both accuracy and robustness on the main task for two of the three datasets. These results highlight two key advantages of CTA, which are supported by findings in recent literature. \cite[Section 4.6]{gontijolopes2022representationrulealloverlapping} demonstrates that trivially mixing the feature vectors of encoders trained under different frameworks yields an accuracy score higher than any of the individual models. CTA extends this property by making the mixing of features a learning objective, explaining the increase in source accuracy after alignment. \cite{shi2022robustunsupervisedrepresentationlearning} shows that self-supervised learning (SSL) is more robust to large distribution shifts than its supervised counterpart as it is less likely to rely on spurious correlations. Unlike previous TTT methods, CTA never exposes its test-time encoder (the self-supervised model) to supervised learning and instead distills the decision boundary of the supervised task via a SSL objective, thereby entirely preserving its intrinsic robustness. This explains the increased robustness of the aligned model on target data.

We ablate with increasing values for $\tau$ in \cref{eq:lalign} and find that performance increases proportionally. For our final experiments, we use $\tau$=0.5, which follows TTT++ \cite{ttt++}. However, the inverse is true for test-time training, where we find that smaller $\tau$ values yield better results. Consequently, our $\tau$ value for both the source training and test-time training is 0.01.

\subsection{Test-time training results}
\label{sec:ttt_results}

Following the evaluation of the alignment phase, we perform test-time training on the target domain. Our method is compared to our ResNet50 baseline, and four recent TTT approaches, namely TTT++ \cite{ttt++}, ClusT3 \cite{hakim2023clust3}, NC-TTT \cite{NCTTT2024} and ReC-TTT \cite{rec-ttt}.

\mypar{CIFAR10-C} \cref{tab:cifar10c_results} shows our results on the CIFAR10-C benchmark. Our approach outperforms all previous methods on all corruption types. Notably, our average performance demonstrates a 4.51$\%$ gain on ReC-TTT \cite{rec-ttt}, the most recent state-of-the-art, and a 21.43$\%$ gain on our baseline. Additionally, our method is on par with NC-TTT for the smallest amount of variability (i.e., $\pm 0.22$), as opposed to $\pm 5.05$, $\pm 2.78$ and $\pm 1.18$ reported by TTT++, ClusT3 and ReC-TTT respectively.

\mypar{CIFAR100-C} \cref{tab:cifar100c_results} presents our results on the CIFAR100-C benchmark. While ClusT3 or ReC-TTT surpass CTA on certain corruptions, such as \emph{Impulse Noise} and \emph{Elastic Transform}, our approach consistently outperforms prior methods across most corruption types. Specifically, it achieves an average improvement of 21.02\% over the baseline and a 3.29\% average gain over ReC-TTT. These results further demonstrate the effectiveness of our model in scenarios with an increased number of classes.

\mypar{TinyImageNet-C} \cref{tab:tinyimagenetc_results} presents our results on the TinyImageNet-C benchmark. While ReC-TTT achieves superior performance on three corruption types (\emph{Gaussian Noise}, \emph{Shot Noise}, and \emph{Impulse Noise}), indicating greater robustness to noise-based perturbations, our approach outperforms prior methods across all other corruption types. Specifically, it achieves an average improvement of 16.31\% over the baseline and a 4.27\% average gain over ReC-TTT. These results further highlight the robustness of our method when applied to a larger dataset with an increased number of classes.

\subsection{Mitigation of conflictual updates}
\label{sec:alignment_testtime}

\begin{table*}[h]
\centering
{\small 
  \setlength{\tabcolsep}{10pt}
  \begin{tabular}{lccccc}
    \toprule
    \multirow[c]{2}{*}{\textbf{Corruption Type}} & \multirow[c]{2}{*}{\textbf{ResNet50}} & {\textbf{ClusT3}}  & {\textbf{NC-TTT}} & {\textbf{ReC-TTT}}  & \multirow[c]{2}{*}{\textbf{CTA} (Ours)} \\[-3.5pt]
    &  & \scriptsize{(ICCV23)} &  \scriptsize{(CVPR24)} &  \scriptsize{(WACV25)} &\\
    \midrule
    Gaussian Noise & \phz6.87 & 32.65 & 31.92 &\textbf{34.87} & 29.65\\
    Shot Noise & \phz9.99 & 34.72 & 34.47 & \textbf{36.60} & 34.39 \\
    Impulse Noise & \phz3.91 & 22.78 & 22.78 & \textbf{26.09} & 22.54 \\
    Defocus Blur & 18.66 & 29.08 & 25.28 & 31.09 & \textbf{39.46} \\
    Glass Blur & 14.06 & 16.26 & 15.67 & 19.59 & \textbf{26.50} \\
    Motion Blur & 35.30 & 43.92 & 43.39 & 45.55 & \textbf{52.18} \\
    Zoom Blur & 35.13 & 41.17 & 40.46 & 42.53 & \textbf{49.26} \\
    Snow & 32.40 & 42.97 & 43.46 & 40.33 & \textbf{44.71} \\
    Frost & 37.37 & 45.32 & 45.51 & 44.59 & \textbf{49.73} \\
    Fog & 14.86 & 37.85 & 37.68 & 33.08 & \textbf{41.05} \\
    Brightness & 33.94 & 51.19 & 50.62 & 48.53 & \textbf{55.47} \\
    Contrast & \phz2.65 & 2.27 & 2.27 & 8.32 & \textbf{19.58} \\
    Elastic Transform & 41.02 & 41.60 & 41.47 & 44.91 & \textbf{50.92}\\
    Pixelate & 40.38 & 37.00 & 39.31 & 52.96 & \textbf{53.95} \\
    JPEG Compression & 55.17 & 50.57 & 50.91 & 53.32 & \textbf{57.02} \\
    \midrule
    Average & 25.45 & 35.29 & 35.01 & 37.49 & \textbf{41.76} \\
    $\Delta$(Ours$-$Other) & \better{16.31}  & \better{6.47} & \better{6.75} & \better{4.27} & --\\
    \bottomrule
  \end{tabular}
  }
  \caption{Accuracy ($\%$) on TinyImagenet-C dataset with level 5 corruption for CTA and recent state-of-the-art methods.}
  \label{tab:tinyimagenetc_results}
\end{table*}

\begin{table}[h]
\begin{center}
\begin{footnotesize}
\def\arraystretch{1}
\resizebox{\linewidth}{!}{
\begin{tabular}{@{}l cc cc@{}}
\toprule
  & \multicolumn{2}{c}{\textbf{Y model}} & \multicolumn{2}{c}{\textbf{CTA (ours)}} \\
    \cmidrule(l{4pt}r{4pt}){2-3}
    \cmidrule(l{4pt}r{4pt}){4-5}
  & Encoder & Projector & Encoder & Projector \\
\midrule
\multicolumn{5}{c}{\textbf{Semantic alignment with supervised task (DBI $\downarrow$)}} \\
\bottomrule
\addlinespace[0.5ex]
No adapt.  & 3.22 & 5.64 & 2.98 & 1.85 \\
20 adapt iter. & 3.43 & 6.11 & 2.86 & 2.09 \\
\midrule
\multicolumn{5}{c}{\textbf{Distance to source classifier embedding-space ($\downarrow$)}} \\
\bottomrule
\addlinespace[0.5ex]
No adapt.  & 83.00  & --     & --     & 192.08 \\
20 adapt iter. & 107.52 & --     & --     & 190.36 \\
\midrule
\end{tabular}
}
\caption{Semantic alignment and distance-to-source results for CTA and Y model. Results are averaged across all CIFAR10-C corruptions for severity 5.}
\label{tab:alignment_results}
\end{footnotesize}
\end{center}
\end{table}

\begin{figure*}[h]
  \centering
  \includegraphics[width=.95\textwidth]{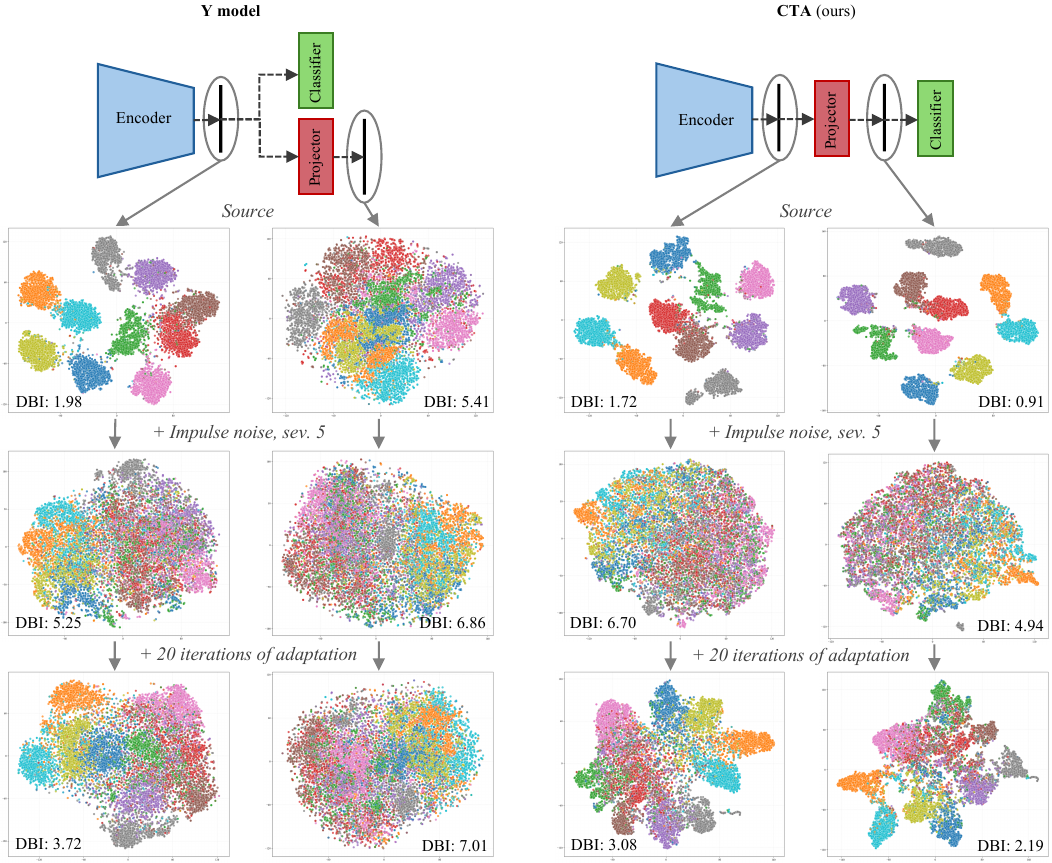}
  \caption{Comparative visualization of domain representations as t-SNE plots between the Y model and CTA for CIFAR10 and CIFAR10-C. The DBI scores are reported within the plot for each representation. The first row shows the source domain, the second column shows the unadapted models with impulse noise corruption of severity 5 and the third column shows the models on the same corruption after 20 iterations of adaptation. The feature extractor of CTA, specifically the encoder and projector, is consistently more robust to domain shift and more semantically aligned than the feature extractor of the Y model.}
  \label{fig:latent_spaces}
\end{figure*}

To assess whether the alignment phase mitigates conflictual updates during test-time training, we compare the embedding space of CTA to that of a standard multi-task TTT architecture (Y model). To this end, we jointly train a classifier and a projector on the same supervised and self-supervised objectives used in CTA, but with both modules operating from a common feature extractor. Following the source training, we perform test-time training by updating only the self-supervised branch. We evaluate two key aspects: \textbf{(1)} the extent to which the self-supervised features produced by the projector are semantically aligned with the supervised task, and \textbf{(2)} whether self-supervised updates at test time move the features closer to the frozen classifier’s learnt embedding space.

To quantify semantic alignment, we use the Davies–Bouldin Index (DBI) with respect to the ground-truth labels. To assess how much the updated features drift from those known to the classifier, we measure the distance between the median class centroids before and after test-time training. For CTA, the reference is the output of the projector $\pi(\cdot)$ whereas for the Y model, it is the output of the encoder.

\cref{fig:latent_spaces} shows the t-SNE plots of the features at different stages of source and test-time training. While the Y model has to bring the shared backbone back into alignment using misaligned representations, CTA's projector yields representations that are both more semantically aligned and more robust than the Y model's, even after 20 adaptation steps. \cref{tab:alignment_results} shows the same results for CIFAR10-C averaged across all corruptions of severity 5. This alignment enables more meaningful self-supervised updates towards the supervised objective. As a result, CTA reduces its distance from the classifier's embedding space during test-time training, while the Y model drifts further away. 

\subsection{Number of iterations needed}
\label{sec:architecture_agnostic}

\begin{figure}[h]
  \centering
  \includegraphics[width=\linewidth]{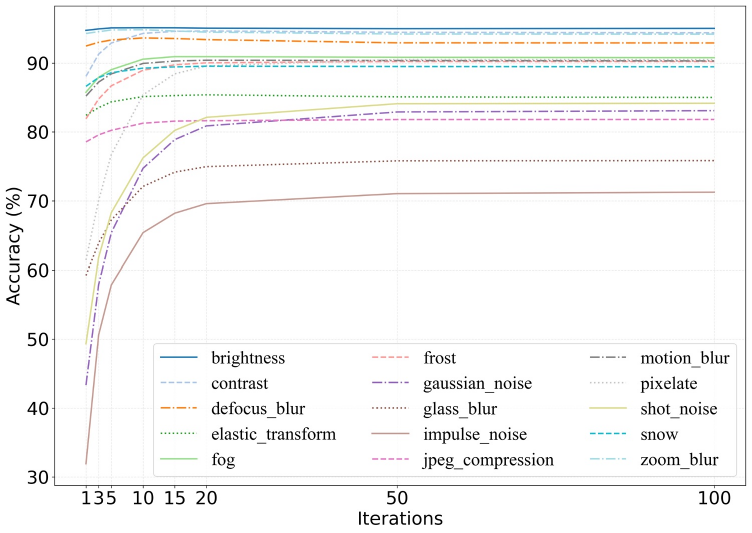}
  \caption{Evolution of accuracy (\%) as a function of the number of iterations for all corruptions in CIFAR-10-C.} 
  \label{fig:adapt_progression_cifar10c}
\end{figure}

An important aspect of test-time training (TTT) is determining the number of iterations required to achieve optimal performance on the target domain. The unsupervised nature of TTT makes it challenging to detect overfitting, as no labeled target domain samples are available for validation. Following previous work \cite{rec-ttt, NCTTT2024}, we report the progression of classification accuracy of CTA on the CIFAR10-C benchmark over iterations of adaptation in \cref{fig:adapt_progression_cifar10c}. Here, an iteration refers to the number of times the model processes each batch of the target dataset. Our results indicate that CTA rapidly adapts to domain shifts, achieving a significant performance improvement within 20 iterations. Furthermore, accuracy remains stable even after reaching optimal performance, suggesting robustness to excessive adaptation.

\subsection{Effect of the supervised encoder}

To study the impact of the supervised encoder during the alignment phase, we conduct an experiment using only a self-supervised model and a classifier. First, we train the self-supervised model \(\pi(g(\cdot))\) using \cref{eq:lcon}. We then freeze the self-supervised model and use it as a feature extractor to train a classifier \(h(\cdot)\). At test-time, we adapt the self-supervised model using the same loss it was trained with. We show the results of this ablation study as CTA-C in \cref{tab:cifar10c_results}. While CTA-C demonstrates an average loss of 4.48$\%$ compared to CTA, it outperforms ReC-TTT with an average gain of 0.02$\%$.

\section{Conclusion}
    \label{sec:conclusion}
    
We propose CTA, an architecture-agnostic test-time training method inspired by recent advances in contrastive learning. Motivated by the known problem of gradient interference as a result of the multi-task training framework of common TTT methods \cite{ttt++, ruder2017overviewmultitasklearningdeep, yu2020gradientsurgerymultitasklearning}, we use contrastive learning to align a self-supervised model with a supervised objective, thereby avoiding conflicting gradient updates and preserving its intrinsic robustness. To this end, we initially train two identical models on their respective tasks. Then, we align their feature distributions using a student-teacher framework, whereby the self-supervised model learns to match the output embeddings of a frozen supervised encoder, while still optimizing its auxiliary task. Despite never being exposed to the cross-entropy loss, the aligned self-supervised model outperforms the supervised encoder in terms of both performance and robustness on the main task. CTA is evaluated on three publicly available datasets and compared with recent state-of-the-art test-time training methods. Across all datasets, CTA consistently demonstrates superior performance.

As our method is architecture-agnostic and applicable to any auxiliary tasks, other experimental settings such as mask-modeling with vision transformers could be explored. Additionally, while our work was under computational restraints, we encourage future work to scale CTA to larger foundation models.

{
    \small
    \bibliographystyle{ieeenat_fullname}
    \bibliography{main}

\begin{thebibliography}{37}
\providecommand{\natexlab}[1]{#1}
\providecommand{\url}[1]{\texttt{#1}}
\expandafter\ifx\csname urlstyle\endcsname\relax
  \providecommand{\doi}[1]{doi: #1}\else
  \providecommand{\doi}{doi: \begingroup \urlstyle{rm}\Url}\fi

\bibitem[Cha et~al.(2021)Cha, Chun, Lee, Cho, Park, Lee, and Park]{cha2021swad}
Junbum Cha, Sanghyuk Chun, Kyungjae Lee, Han-Cheol Cho, Seunghyun Park, Yunsung Lee, and Sungrae Park.
\newblock Swad: Domain generalization by seeking flat minima.
\newblock \emph{Advances in Neural Information Processing Systems}, 34:\penalty0 22405--22418, 2021.

\bibitem[Chen et~al.(2020)Chen, Kornblith, Norouzi, and Hinton]{SimCLR}
Ting Chen, Simon Kornblith, Mohammad Norouzi, and Geoffrey Hinton.
\newblock A simple framework for contrastive learning of visual representations.
\newblock \emph{arXiv preprint arXiv:2002.05709}, 2020.

\bibitem[Colussi et~al.(2024{\natexlab{a}})Colussi, Mascetti, Dolz, and Desrosiers]{colussi2024rec}
Marco Colussi, Sergio Mascetti, Jose Dolz, and Christian Desrosiers.
\newblock {ReC-TTT}: Contrastive feature reconstruction for test-time training.
\newblock \emph{arXiv preprint arXiv:2411.17869}, 2024{\natexlab{a}}.

\bibitem[Colussi et~al.(2024{\natexlab{b}})Colussi, Mascetti, Dolz, and Desrosiers]{rec-ttt}
Marco Colussi, Sergio Mascetti, Jose Dolz, and Christian Desrosiers.
\newblock Rec-ttt: Contrastive feature reconstruction for test-time training, 2024{\natexlab{b}}.

\bibitem[Deng et~al.(2009)Deng, Dong, Socher, Li, Li, and Fei-Fei]{imagenet}
Jia Deng, Wei Dong, Richard Socher, Li-Jia Li, Kai Li, and Li Fei-Fei.
\newblock Imagenet: A large-scale hierarchical image database.
\newblock In \emph{2009 IEEE Conference on Computer Vision and Pattern Recognition}, pages 248--255, 2009.

\bibitem[Gandelsman et~al.(2022)Gandelsman, Sun, Chen, and Efros]{gandelsman2022test}
Yossi Gandelsman, Yu Sun, Xinlei Chen, and Alexei Efros.
\newblock Test-time training with masked autoencoders.
\newblock \emph{Advances in Neural Information Processing Systems}, 35:\penalty0 29374--29385, 2022.

\bibitem[Gontijo-Lopes et~al.(2022)Gontijo-Lopes, Dauphin, and Cubuk]{gontijolopes2022representationrulealloverlapping}
Raphael Gontijo-Lopes, Yann Dauphin, and Ekin~D. Cubuk.
\newblock No one representation to rule them all: Overlapping features of training methods, 2022.

\bibitem[Hakim et~al.(2023)Hakim, Osowiechi, Noori, Cheraghalikhani, Bahri, Ben~Ayed, and Desrosiers]{hakim2023clust3}
Gustavo A~Vargas Hakim, David Osowiechi, Mehrdad Noori, Milad Cheraghalikhani, Ali Bahri, Ismail Ben~Ayed, and Christian Desrosiers.
\newblock Clust3: Information invariant test-time training.
\newblock In \emph{Proceedings of the IEEE/CVF International Conference on Computer Vision}, pages 6136--6145, 2023.

\bibitem[He et~al.(2015)He, Zhang, Ren, and Sun]{He2015}
Kaiming He, Xiangyu Zhang, Shaoqing Ren, and Jian Sun.
\newblock Deep residual learning for image recognition.
\newblock \emph{arXiv preprint arXiv:1512.03385}, 2015.

\bibitem[Hendrycks and Dietterich(2019)]{hendrycks2019robustness}
Dan Hendrycks and Thomas Dietterich.
\newblock Benchmarking neural network robustness to common corruptions and perturbations.
\newblock \emph{Proceedings of the International Conference on Learning Representations}, 2019.

\bibitem[Hinton et~al.(2015)Hinton, Vinyals, and Dean]{hinton2015distillingknowledgeneuralnetwork}
Geoffrey Hinton, Oriol Vinyals, and Jeff Dean.
\newblock Distilling the knowledge in a neural network, 2015.

\bibitem[Krizhevsky(2009)]{CIFAR10}
Alex Krizhevsky.
\newblock Learning multiple layers of features from tiny images.
\newblock Technical report, University of Toronto, 2009.

\bibitem[Le and Yang(2015)]{TinyImagenet}
Ya Le and Xuan Yang.
\newblock Tiny imagenet visual recognition challenge, 2015.

\bibitem[Liang et~al.(2020)Liang, Hu, and Feng]{liang2020we}
Jian Liang, Dapeng Hu, and Jiashi Feng.
\newblock Do we really need to access the source data? source hypothesis transfer for unsupervised domain adaptation.
\newblock In \emph{International conference on machine learning}, pages 6028--6039. PMLR, 2020.

\bibitem[Liu et~al.(2021{\natexlab{a}})Liu, Kothari, Van~Delft, Bellot-Gurlet, Mordan, and Alahi]{liu2021ttt++}
Yuejiang Liu, Parth Kothari, Bastien Van~Delft, Baptiste Bellot-Gurlet, Taylor Mordan, and Alexandre Alahi.
\newblock Ttt++: When does self-supervised test-time training fail or thrive?
\newblock \emph{Advances in Neural Information Processing Systems}, 34:\penalty0 21808--21820, 2021{\natexlab{a}}.

\bibitem[Liu et~al.(2021{\natexlab{b}})Liu, Kothari, van Delft, Bellot-Gurlet, Mordan, and Alahi]{ttt++}
Yuejiang Liu, Parth Kothari, Bastien~Germain van Delft, Baptiste Bellot-Gurlet, Taylor Mordan, and Alexandre Alahi.
\newblock Ttt++: When does self-supervised test-time training fail or thrive?
\newblock In \emph{Thirty-Fifth Conference on Neural Information Processing Systems}, 2021{\natexlab{b}}.

\bibitem[Miller et~al.(2021)Miller, Taori, Raghunathan, Sagawa, Koh, Shankar, Liang, Carmon, and Schmidt]{miller2021accuracy}
John~P Miller, Rohan Taori, Aditi Raghunathan, Shiori Sagawa, Pang~Wei Koh, Vaishaal Shankar, Percy Liang, Yair Carmon, and Ludwig Schmidt.
\newblock Accuracy on the line: on the strong correlation between out-of-distribution and in-distribution generalization.
\newblock In \emph{International Conference on Machine Learning}, pages 7721--7735. PMLR, 2021.

\bibitem[Nado et~al.(2020)Nado, Padhy, Sculley, D'Amour, Lakshminarayanan, and Snoek]{nado2020evaluating}
Zachary Nado, Shreyas Padhy, D Sculley, Alexander D'Amour, Balaji Lakshminarayanan, and Jasper Snoek.
\newblock Evaluating prediction-time batch normalization for robustness under covariate shift.
\newblock \emph{arXiv preprint arXiv:2006.10963}, 2020.

\bibitem[Nguyen et~al.(2023)Nguyen, Nguyen-Tang, Lim, and Torr]{nguyen2023tipi}
A~Tuan Nguyen, Thanh Nguyen-Tang, Ser-Nam Lim, and Philip~HS Torr.
\newblock Tipi: Test time adaptation with transformation invariance.
\newblock In \emph{Proceedings of the IEEE/CVF Conference on Computer Vision and Pattern Recognition}, pages 24162--24171, 2023.

\bibitem[Osowiechi et~al.(2023)Osowiechi, Hakim, Noori, Cheraghalikhani, Ben~Ayed, and Desrosiers]{osowiechi2023tttflow}
David Osowiechi, Gustavo A~Vargas Hakim, Mehrdad Noori, Milad Cheraghalikhani, Ismail Ben~Ayed, and Christian Desrosiers.
\newblock Tttflow: Unsupervised test-time training with normalizing flow.
\newblock In \emph{Proceedings of the IEEE/CVF Winter Conference on Applications of Computer Vision}, pages 2126--2134, 2023.

\bibitem[Osowiechi et~al.(2024{\natexlab{a}})Osowiechi, Hakim, Noori, Cheraghalikhani, Bahri, Yazdanpanah, Ayed, and Desrosiers]{NCTTT2024}
David Osowiechi, Gustavo A.~Vargas Hakim, Mehrdad Noori, Milad Cheraghalikhani, Ali Bahri, Moslem Yazdanpanah, Ismail~Ben Ayed, and Christian Desrosiers.
\newblock Nc-ttt: A noise constrastive approach for test-time training.
\newblock In \emph{***}, 2024{\natexlab{a}}.

\bibitem[Osowiechi et~al.(2024{\natexlab{b}})Osowiechi, Hakim, Noori, Cheraghalikhani, Bahri, Yazdanpanah, Ben~Ayed, and Desrosiers]{osowiechi2024nc}
David Osowiechi, Gustavo A~Vargas Hakim, Mehrdad Noori, Milad Cheraghalikhani, Ali Bahri, Moslem Yazdanpanah, Ismail Ben~Ayed, and Christian Desrosiers.
\newblock Nc-ttt: A noise constrastive approach for test-time training.
\newblock In \emph{Proceedings of the IEEE/CVF Conference on Computer Vision and Pattern Recognition}, pages 6078--6086, 2024{\natexlab{b}}.

\bibitem[Osowiechi et~al.(2024{\natexlab{c}})Osowiechi, Noori, Vargas~Hakim, Yazdanpanah, Bahri, Cheraghalikhani, Dastani, Beizaee, Ayed, and Desrosiers]{watt}
David Osowiechi, Mehrdad Noori, Gustavo Vargas~Hakim, Moslem Yazdanpanah, Ali Bahri, Milad Cheraghalikhani, Sahar Dastani, Farzad Beizaee, Ismail Ayed, and Christian Desrosiers.
\newblock Watt: Weight average test time adaptation of clip.
\newblock In \emph{Advances in Neural Information Processing Systems}, pages 48015--48044, 2024{\natexlab{c}}.

\bibitem[Radford et~al.(2021)Radford, Kim, Hallacy, Ramesh, Goh, Agarwal, Sastry, Askell, Mishkin, Clark, et~al.]{radford2021learning}
Alec Radford, Jong~Wook Kim, Chris Hallacy, Aditya Ramesh, Gabriel Goh, Sandhini Agarwal, Girish Sastry, Amanda Askell, Pamela Mishkin, Jack Clark, et~al.
\newblock Learning transferable visual models from natural language supervision.
\newblock In \emph{International conference on machine learning}, pages 8748--8763. PMLR, 2021.

\bibitem[Ruder(2017)]{ruder2017overviewmultitasklearningdeep}
Sebastian Ruder.
\newblock An overview of multi-task learning in deep neural networks, 2017.

\bibitem[Saenko et~al.(2010)Saenko, Kulis, Fritz, and Darrell]{saenko2010adapting}
Kate Saenko, Brian Kulis, Mario Fritz, and Trevor Darrell.
\newblock Adapting visual category models to new domains.
\newblock In \emph{Computer Vision--ECCV 2010: 11th European Conference on Computer Vision, Heraklion, Crete, Greece, September 5-11, 2010, Proceedings, Part IV 11}, pages 213--226. Springer, 2010.

\bibitem[Shi et~al.(2022)Shi, Daunhawer, Vogt, Torr, and Sanyal]{shi2022robustunsupervisedrepresentationlearning}
Yuge Shi, Imant Daunhawer, Julia~E. Vogt, Philip H.~S. Torr, and Amartya Sanyal.
\newblock How robust is unsupervised representation learning to distribution shift?, 2022.

\bibitem[Shu et~al.(2022)Shu, Nie, Huang, Yu, Goldstein, Anandkumar, and Xiao]{tpt}
Manli Shu, Weili Nie, De-An Huang, Zhiding Yu, Tom Goldstein, Anima Anandkumar, and Chaowei Xiao.
\newblock Test-time prompt tuning for zero-shot generalization in vision-language models.
\newblock In \emph{Advances in Neural Information Processing Systems}, pages 14274--14289. Curran Associates, Inc., 2022.

\bibitem[Sun et~al.(2020{\natexlab{a}})Sun, Wang, Liu, Miller, Efros, and Hardt]{sun2020test}
Yu Sun, Xiaolong Wang, Zhuang Liu, John Miller, Alexei Efros, and Moritz Hardt.
\newblock Test-time training with self-supervision for generalization under distribution shifts.
\newblock In \emph{International conference on machine learning}, pages 9229--9248. PMLR, 2020{\natexlab{a}}.

\bibitem[Sun et~al.(2020{\natexlab{b}})Sun, Wang, Zhuang, Miller, Hardt, and Efros]{sun19ttt}
Yu Sun, Xiaolong Wang, Liu Zhuang, John Miller, Moritz Hardt, and Alexei~A. Efros.
\newblock Test-time training with self-supervision for generalization under distribution shifts.
\newblock In \emph{ICML}, 2020{\natexlab{b}}.

\bibitem[Torralba and Efros(2011)]{torralba2011unbiased}
Antonio Torralba and Alexei~A Efros.
\newblock Unbiased look at dataset bias.
\newblock In \emph{CVPR 2011}, pages 1521--1528. IEEE, 2011.

\bibitem[{Vargas Hakim} et~al.(2024){Vargas Hakim}, Osowiechi, Noori, Cheraghalikhani, Bahri, Yazdanpanah, Ayed, and Desrosiers]{clipartt}
Gustavo~Adolfo {Vargas Hakim}, David Osowiechi, Mehrdad Noori, Milad Cheraghalikhani, Ali Bahri, Moslem Yazdanpanah, Ismail~Ben Ayed, and Christian Desrosiers.
\newblock Clipartt: Light-weight adaptation of clip to new domains at test time, 2024.

\bibitem[Venkateswara et~al.(2017)Venkateswara, Eusebio, Chakraborty, and Panchanathan]{venkateswara2017deep}
Hemanth Venkateswara, Jose Eusebio, Shayok Chakraborty, and Sethuraman Panchanathan.
\newblock Deep hashing network for unsupervised domain adaptation.
\newblock In \emph{Proceedings of the IEEE conference on computer vision and pattern recognition}, pages 5018--5027, 2017.

\bibitem[Wang et~al.(2020)Wang, Shelhamer, Liu, Olshausen, and Darrell]{wang2020tent}
Dequan Wang, Evan Shelhamer, Shaoteng Liu, Bruno Olshausen, and Trevor Darrell.
\newblock Tent: Fully test-time adaptation by entropy minimization.
\newblock \emph{arXiv preprint arXiv:2006.10726}, 2020.

\bibitem[Yu et~al.(2020)Yu, Kumar, Gupta, Levine, Hausman, and Finn]{yu2020gradientsurgerymultitasklearning}
Tianhe Yu, Saurabh Kumar, Abhishek Gupta, Sergey Levine, Karol Hausman, and Chelsea Finn.
\newblock Gradient surgery for multi-task learning, 2020.

\bibitem[Zhou et~al.(2021)Zhou, Yang, Qiao, and Xiang]{zhou2021domain}
Kaiyang Zhou, Yongxin Yang, Yu Qiao, and Tao Xiang.
\newblock Domain generalization with mixstyle.
\newblock \emph{arXiv preprint arXiv:2104.02008}, 2021.

\bibitem[Zhou et~al.(2022)Zhou, Liu, Qiao, Xiang, and Loy]{zhou2022domain}
Kaiyang Zhou, Ziwei Liu, Yu Qiao, Tao Xiang, and Chen~Change Loy.
\newblock Domain generalization: A survey.
\newblock \emph{IEEE Transactions on Pattern Analysis and Machine Intelligence}, 2022.

\end{thebibliography}
}

\end{document}